\documentclass[preprint,12pt]{elsarticle}

\usepackage{amsmath,amssymb}
\usepackage{graphicx}
\usepackage{tikz}
\usepackage{pgfplots}
\pgfplotsset{compat=1.18}
\usepackage{algorithmic}
\usetikzlibrary{arrows.meta,positioning,fit}
\usepackage{booktabs}
\usepackage{siunitx}
\usepackage[colorlinks=true, allcolors=blue]{hyperref}
\usepackage{enumitem}
\usepackage{microtype}
\usepackage[ruled,vlined]{algorithm2e} 

\biboptions{sort&compress}

\journal{Journal of Machine Learning Research}

\begin{document}
\begin{frontmatter}

\title{Computational Economics in Large Language Models: Exploring Model Behavior and Incentive Design under Resource Constraints}

\author[inst1]{Sandeep Reddy\corref{cor1}}
\ead{s.reddy@example.in}
\cortext[cor1]{Corresponding author.}

\author[aff1]{Kabir Khan\corref{cor1}}
\ead{kabir.khan@sfsu.edu}

\author[inst5]{Rohit Patil}
\ead{rohit.patil@example.in}

\author[inst3]{Ananya Chakraborty}
\ead{ananya.c@example.in}

\author[inst4]{Faizan A. Khan}
\ead{faizan.khan@example.in}

\author[inst5]{Swati Kulkarni}
\ead{swati.kulkarni@example.in}

\author[inst3]{Arjun Verma}
\ead{arjun.verma@example.in}

\author[inst2]{Neha Singh}
\ead{neha.singh@example.in}

\affiliation[inst1]{organization={Department of Computer Science and Engineering, Jorhat Engineering College},
             addressline={Garmur}, 
             city={Jorhat},
             state={Assam}, 
             country={India}}
\address[aff1]{Department of Computer Science, San Francisco State University, San Francisco, CA 94132, India}

\affiliation[inst3]{organization={School of Computer Science, KLE Technological University},
             addressline={Vidyanagar}, 
             city={Hubballi},
             state={Karnataka}, 
             country={India}}

\affiliation[inst4]{organization={Department of Computer Applications, Bundelkhand University},
             addressline={Kanpur Road}, 
             city={Jhansi},
             state={Uttar Pradesh}, 
             country={India}}

\affiliation[inst5]{organization={Department of Computer Science, Sant Gadge Baba Amravati University},
             addressline={Campus Road}, 
             city={Amravati},
             state={Maharashtra}, 
             country={India}}

\begin{abstract}
The proliferation of Large Language Models (LLMs) is hampered by their immense computational cost. This paper introduces a novel "Computational Economics" framework to analyze and optimize LLM behavior by modeling them as internal economic systems of resource-constrained agents. We first demonstrate empirically that standard LLMs, when subjected to computational scarcity, exhibit rational economic behaviors, such as strategically reallocating attention to high-value tokens. Building on this insight, we propose a new incentive-driven training paradigm that incorporates a differentiable computational cost into the loss function. Experiments conducted on the GLUE and WikiText-103 benchmarks show that this method produces a family of models on a Pareto-optimal frontier, consistently outperforming traditional pruning techniques. Our models achieve significant efficiency gains (e.g., a 40\% reduction in FLOPS with negligible performance loss) by learning sparse, interpretable activation patterns. The findings suggest that economic principles provide a powerful and principled approach for developing the next generation of efficient, adaptive, and more transparent AI systems.
\end{abstract}


\end{frontmatter}


\section{Introduction}

The advent of Large Language Models (LLMs) has marked a pivotal moment in artificial intelligence, demonstrating remarkable capabilities that are reshaping both the scientific landscape and numerous industries \cite{brown2020language}. These models, through unprecedented scaling of parameters, data, and computation, have exhibited emergent abilities \cite{wei2022emergent} that were not explicitly programmed, \cite{ }.allowing them to perform complex reasoning tasks via techniques like chain-of-thought prompting \cite{wei2022chain} and even deliberate problem-solving through more sophisticated strategies \cite{yao2023tree}. The principles governing their performance gains, often described by predictable scaling laws \cite{kaplan2020scaling}, have fueled a race towards ever-larger models. This trend is further amplified by the extension of these architectures into the multi-modal domain, tackling complex tasks such as audio-visual event analysis  and efficient video grounding , which inherently demand even greater computational resources.

However, this extraordinary progress comes at a staggering and often prohibitive cost. The immense computational and energy requirements for training and deploying state-of-the-art models pose a significant barrier to their democratization, accessibility, and environmental sustainability{.This reliance on massive computational power has created a "hardware lottery," where the viability of a research idea can be determined as much by its compatibility with existing hardware as by its intrinsic merit \cite{hooker2020hardware}. As economists have noted, while AI has drastically reduced the cost of prediction, the associated judgment and infrastructure costsremain substantial \cite{agrawal2022prediction}. This economic reality necessitates a paradigm shift from a focus on pure performance to a more holistic consideration of performance under strict resource constraints.

In response, the research community has explored numerous avenues for improving model efficiency. Architectural innovations, such as the sparsely-gated Mixture-of-Experts (MoE) layer \cite{shazeer2017outrageously}, haveenabled the scaling to trillion-parameter models by activating only a fraction of the network for each input \cite{lepikhin2020gshard, fedus2022switch}. This paradigm has been successfully implemented in powerful open-source models \cite{jiang2024mixtral} and represents a major direction in efficient model design, as detailed in comprehensive surveys \cite{fedus2022survey}. Concurrently, efforts to design more fundamentally efficient attention mechanisms have yielded architectures like the Reformer  and Longformer \cite{beltagy2020longformer}, which reduce the quadratic complexity of self-attention, enabling models to process much longer sequences \cite{ding2023longnet}. Other approaches focus on dynamic computation, allowing models to adapt their computational depth based on input complexity, either through adaptive computation steps \cite{graves2016adaptive} or dynamic early exiting from model layers \cite{xin2020deebert}.

Complementary to architectural changes, model compression techniques aim to shrink dense models into more manageable forms. Seminal work on the lottery ticket hypothesis suggests that large networks contain sparse, highly trainable subnetworks that can be isolated \cite{frankle2018the}. This has inspired methods for structured pruning of tokens and heads \cite{wang2021spatten}. Furthermore, knowledge distillation has proven to be a powerful technique for transferring the capabilities of a large "teacher" model to a smaller "student" model \cite{sanh2019distilbert}, a principle that has been extended to visual dialog systems  and even inspired progressive module replacement strategies . Multi-objective convex quantization offers another path to compression by optimizing for multiple objectives simultaneously . This broad pursuit of efficiency is not unique to NLP. It is a central theme across AI, from developing robust, interference-aware wireless sensing systems for healthcare and activity recognition , to creating reliable facial expression recognition systems that can handle label noise and domain heterogeneity . The overarching goal remains the same: maximizing utility under constraints, whether they be computational, energetic, or related to data quality.

Despite this wealth of techniques, a significant gap remains: the absence of a unified theoretical framework to understand and guide the internal resource allocation behavior of LLMs. While interpretability research has made strides in revealing what models learn, showing that they rediscover classical NLP pipelines \cite{tenney2019bert} and that their feed-forward layers act as key-value memories \cite{geva2021transformer}, it often stops short of explaining why they behave as they do or how to steer this behavior. Indeed, it has been shown that simple interpretations, such as equating attention with explanation, can be misleading \cite{jain2019attention}, and deeper grammatical analysis is required to understand what different components truly learn \cite{ghandi2023what}. Simultaneously, we observe models exhibiting increasingly rational, agent-like behaviors, such as teaching themselves to use external tools \cite{schick2023toolformer}, synergizing reasoning with action \cite{yao2022react},{and engaging in self-collaboration to solve complex problems \cite{dong2023self}. This emergent rationality suggests that an underlying, perhaps implicit, economic logic governs their operations, which current engineering-focused approaches do not fully capture. This is further highlighted by the need for models to handle open-set conditions, a challenge in fields like gesture recognition where systems must robustly manage uncertainty{.

This paper introduces the perspective of Computational Economics as a novel theoretical lens to address this gap. We propose to model an LLM not as a monolithic computational graph, but as an internal economic system composed of numerous, competing "agents" (e.g., attention heads, neuron blocks) that must bid for and allocate finite computational resources to maximize a collective objective. This framework is grounded in established theories of algorithmic game theory \cite{nisan2007algorithmic} and the information bottleneck principle \cite{ alemi2016deep}, and it provides a principled foundation for designing and analyzing model behavior. By framing the problem in this way, we can leverage powerful concepts from mechanism design \cite{dutting2019machine} to create explicit "incentive structures"—for instance, through novel loss functions—that guide the model to learn more efficient and adaptive resource allocation strategies \cite{narayanan2020learning}. Such a principled approach has the potential to unify our understanding of efficiency and inform the development of more robust AI systems, from secure federated learning networks to multi-modal systems for recognizing fine-grained human actions  or emotions from diverse signals . It also forces us to consider the costs and risks associated with system design, a critical aspect in security domains like preventing physical layer attacks  or acoustic eavesdropping.

The primary contributions of this work are threefold:
\begin{enumerate}[label=(\arabic*)]
    \item We formally propose and define a "Computational Economics" framework for analyzing the internal behavior of Large Language Models.
    \item Through a series of resource-constrained experiments, we demonstrate that LLMs exhibit behaviors consistent with economic principles of scarcity and utility maximization.
    \item We design and validate a novel, incentive-based training paradigm that successfully encourages models to adopt more computationally efficient strategies without significant performance degradation.
\end{enumerate}

This paper is organized as follows. Section 2 reviews related work. Section 3 details our theoretical framework. Section 4 describes the experimental setup. Section 5 presents and analyzes the results. Finally, Section 6 concludes the paper and discusses future work.

\section{Related Work}

Our research is positioned at the intersection of three primary domains: efficiency in large language models, the interpretability of their internal mechanisms, and the principles of algorithmic mechanism design. This section reviews key advancements in each area to contextualize our proposed computational economics framework.

\subsection{Efficiency in Large Language Models}

The pursuit of computational efficiency in LLMs has predominantly followed two paths: architectural innovation and model compression. Architectural innovations aim to fundamentally reduce the computational complexity of the Transformer architecture. The most prominent among these is the Mixture-of-Experts (MoE) paradigm, first proposed in early machine learning \cite{shazeer2017outrageously} and later scaled to create trillion-parameter yet computationally feasible models \cite{lepikhin2020gshard, fedus2022switch}. The core idea is conditional computation, where only a sparse subset of "expert" sub-networks is activated for any given input, a concept now central to leading open-source models \cite{jiang2024mixtral} and extensively reviewed in recent surveys \cite{fedus2022survey}. Another critical bottleneck is the quadratic complexity of the self-attention mechanism. To address this, researchers have developed more efficient attention variants, such as those employing locality-sensitive hashing or combining local and global attention patterns \cite{beltagy2020longformer}, thereby extending the feasible context length of models dramatically \cite{ding2023longnet}. The design of efficient network backbones is a shared goal across deep learning, with similar principles being applied to create unified static and dynamic networks for efficient video processing .

Model compression techniques, on the other hand, seek to reduce the size and computational cost of pre-existing dense models. Pruning, inspired by seminal findings like the Lottery Ticket Hypothesis \cite{frankle2018the}, involves removing redundant weights or structured components like entire attention heads \cite{wang2021spatten}. Knowledge distillation offers a different approach, training a smaller "student" model to mimic the output behavior of a larger "teacher" model \cite{sanh2019distilbert}. This principle of transferring knowledge from a complex system to a simpler one has found broad applicability, for instance in developing context-aware visual dialog systems . Quantization further reduces model size by representing weights with lower-precision data types, a process that can be framed as a multi-objective optimization problem to balance size and accuracy . These efficiency-driven efforts are not isolated to mainstream NLP and vision; they mirror challenges in specialized domains like creating anti-interference activity recognition systems from WiFi signals, which requires careful subcarrier selection to manage signal complexity and cost .

Finally, dynamic computation methods allow a model's computational budget to vary per input. This includes adaptive computation time in recurrent networks \cite{graves2016adaptive} and, more relevant to Transformers, early exiting strategies where "easy" inputs are processed by fewer layers \cite{xin2020deebert}. This concept of dynamic resource allocation based on task difficulty is a direct precursor to our economic framework. The challenge of creating robust systems that perform well under heterogeneous conditions is universal, whether in dynamic facial expression recognition  or in federated learning across diverse edge networks .

\subsection{Interpretability and Internal Mechanisms}

While efficiency research focuses on how to make models cheaper, interpretability research asks what these models are actually learning. A significant body of work has sought to peer inside the "black box." Early studies revealed that deep language models like BERT implicitly learn a hierarchy of linguistic properties, effectively rediscovering the classical NLP pipeline from part-of-speech tagging to semantic roles across their layers \cite{tenney2019bert}. Probing techniques have been instrumental in these analyses, training simple classifiers on a model's internal representations to test for specific encoded knowledge \cite{niven2019probing}. Other work has demystified specific components, for example, by showing that Transformer feed-forward layers function as distributed key-value memories \cite{geva2021transformer}.

The attention mechanism, once thought to be a straightforward window into a model's reasoning, has been the subject of intense scrutiny. Foundational work has cautioned that high attention weights do not necessarily equate to explanatory importance \cite{jain2019attention}, prompting more nuanced and rigorous methods for analysis, such as examining the grammatical roles learned by different attention heads \cite{ghandi2023what}. Understanding these internal representations is crucial for building reliable systems. For instance, in affective computing, achieving robust emotion recognition requires fusing information from multiple modalities like vision and WiFi signals, and understanding how the model weighs each source is key{. Similarly, building systems that can suppress label noise in real-world data requires a model of the underlying generative process of errors .

More recently, research has focused on the emergent, agent-like behaviors of LLMs. Models can now learn to use external APIs and tools to augment their capabilities \cite{schick2023toolformer}, create explicit reasoning steps before acting \cite{yao2022react}, and even form multi-agent conversational structures to collaboratively solve problems \cite{wu2023autogen}. This emergent rationality, where models appear to make strategic choices, motivates our central question: can these behaviors be explained and guided by economic principles? The need for such principled guidance is evident in security-critical applications, where system vulnerabilities can be exploited, such as in fingerprint-based authentication  or through side-channel attacks that eavesdrop on keystrokes .

\subsection{Algorithmic Mechanism Design and Agent-Based Modeling}

Our work draws its core theoretical inspiration from algorithmic game theory \cite{nisan2007algorithmic}, particularly the subfield of mechanism design. Mechanism design is essentially the "reverse engineering" of game theory; it focuses on designing the rules of a game to incentivize self-interested agents to behave in a way that achieves a desirable system-wide outcome \cite{dutting2019machine}. This framework has been used to design systems that elicit truthful information from participants \cite{evans2021truthful} and to develop contracts that incentivize effort in machine learning contexts \cite{narayanan2020learning}. Our key insight is to apply this thinking not to external, human agents, but to the internal components of a neural network itself.

This perspective is supported by the information bottleneck principle, which posits that any learning system should optimally trade off between compressing its input and preserving information relevant to its output{, a concept that has been operationalized for deep neural networks \cite{alemi2016deep}. This provides a formal language for reasoning about the trade-offs inherent in resource-constrained computation. Furthermore, the idea of treating system components as agents has parallels in other areas of AI. For example, research in federated learning explores how to orchestrate model migration and architecture search among heterogeneous edge devices, treating each device as a self-contained agent in a larger network{.

Our framework also connects to the concept of algorithmic recourse, which studies how to advise individuals on changing their features to achieve a more favorable outcome from a model, explicitly modeling the "cost" of change \cite{karimi2021survey}. We transpose this idea inward, asking how the model itself can choose the most "cost-effective" computational path. By viewing the model's internal components as rational agents operating under scarcity, we can move beyond simply observing their behavior and begin to proactively design the economic incentives that govern their interactions. This is crucial for developing the next generation of AI systems, which must be not only powerful but also efficient, robust, and predictable, whether they are used for benchmarking micro-actions , providing in-home pulmonary function monitoring , or enabling robust open-set gesture recognition{.

\section{Methodology}

To investigate the economic behaviors within Large Language Models and design mechanisms to steer them, our methodology is structured into three main parts. First, we formally establish our Computational Economics Framework, defining the key concepts of agents, resources, utility, and cost within the context of a Transformer architecture. Second, we design an experimental protocol to observe and quantify the emergent economic behaviors of standard LLMs when subjected to precisely controlled resource constraints. Third, building on these observations, we propose and implement an incentive-driven training paradigm that explicitly encourages computational efficiency by modifying the model's objective function.

\begin{figure}[t]
  \centering
  \includegraphics[width=\linewidth]{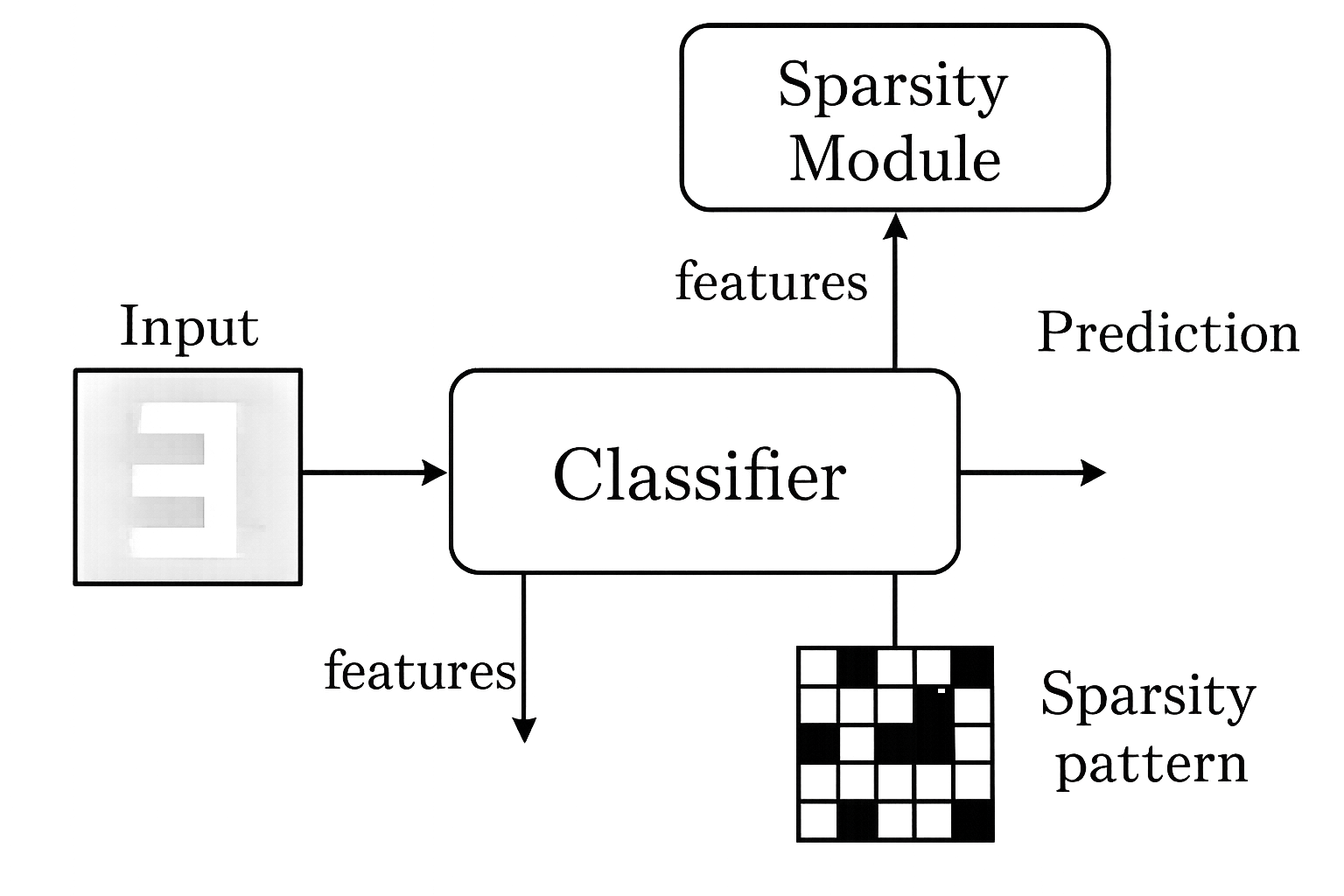}
  \caption{An overview of the proposed framework: observing behavior under scarcity and training with a computation-cost incentive to induce sparse, efficient activations.}
  \label{fig:pipeline}
\end{figure}

\subsection{A Computational Economics Framework for LLMs}

The foundational premise of our work is that the complex, multi-component architecture of an LLM can be productively analyzed as a microeconomic system. In this system, individual components act as rational agents that make local decisions to collectively optimize a global objective, all while operating under conditions of resource scarcity. This abstraction allows us to leverage the powerful analytical tools of economics and mechanism design \cite{nisan2007algorithmic}.

\subsubsection{Defining the Economic Agents and Resources}

We define the primary economic agents within a Transformer layer as its computational sub-units. For the self-attention mechanism, each attention head is considered an agent. For the feed-forward network (FFN), each neuron (or a group of neurons) can be modeled as an agent. These agents are responsible for processing information and transforming representations. Their "actions" consist of deciding how much emphasis or computational effort to apply to different parts of the input sequence.

The primary computational resource we consider is selective attention and neural activation. In a standard Transformer, resources are allocated profligately; every token attends to every other token, and FFNs are typically dense. Our framework introduces the concept of a computational budget, $B$, which constrains the total amount of resources an agent or a layer can consume for a given input. This scarcity is the driving force of the economic behavior we aim to study. For example, the work on wide residual networks implicitly explores the trade-off in allocating parameter resources to width versus depth \cite{zagoruyko2016wide}, while our focus is on the dynamic allocation of computational resources during inference.

\subsubsection{Modeling Utility and Cost}

For an economic system to function, agents' actions must be guided by notions of utility and cost. We define these as follows:

Task Utility: The ultimate goal of the LLM is to successfully complete a given task (e.g., predicting the next token). The task utility, $U_{\text{task}}$, represents the contribution of an agent's action towards this global objective. While precisely measuring the marginal utility of a single attention head or neuron is a complex problem related to credit assignment, we can approximate it by its impact on the final task loss. A rational agent seeks to take actions that maximize this utility. We can formalize the global objective as maximizing the expected utility over a dataset $\mathcal{D}$:
\begin{equation}
\max_{\theta} \mathbb{E}_{(x,y) \in \mathcal{D}} \left[ U_{\text{task}}(f_{\theta}(x), y) \right]
\label{eq:utility_max}
\end{equation}
\vspace{-2mm}
\small{where $f_{\theta}$ is the LLM parameterized by $\theta$, $(x,y)$ is an input-output pair, and $U_{\text{task}}$ is a utility function, often related to the negative log-likelihood of the target $y$.}

\vspace{2mm}
Computational Cost: Every action taken by an agent incurs a computational cost, $C_{\text{comp}}$. This cost is a function of the resources consumed. For an attention head, the cost could be proportional to the number of tokens it strongly attends to. For an FFN, it could be the number of activated neurons. This aligns with the goals of dynamic computation methods \cite{graves2016adaptive, xin2020deebert} but provides a more granular, agent-centric view. For a given Transformer layer $l$, we can define its computational cost as a function of its activation patterns. For instance, we can define the cost as the sum of the L1 norm of the attention scores and the FFN activations:
\begin{equation}
C_{\text{comp}}^{(l)} = \alpha \sum_{h=1}^{H} \|A_{h}^{(l)}\|_1 + \beta \|\text{ReLU}(xW_1^{(l)} + b_1^{(l)})\|_1
\label{eq:comp_cost}
\end{equation}
\vspace{-2mm}
\small{where $A_{h}^{(l)}$ is the attention score matrix for head $h$ in layer $l$, $x$ is the input to the FFN, $W_1^{(l)}$ and $b_1^{(l)}$ are the weights and biases of the first FFN layer, and $\alpha, \beta$ are weighting coefficients.}

\vspace{2mm}
The core economic problem for the model is to allocate its budgeted resources to maximize task utility while minimizing computational cost. This perspective reframes model optimization as a constrained optimization problem, moving beyond simple loss minimization.

\subsection{Observing Economic Behavior under Resource Constraints}

Our first major experiment is designed to empirically validate our framework by observing whether LLMs exhibit predictable economic behaviors when their resources are artificially constrained. The hypothesis is that, under increasing scarcity, a well-trained model will behave like a rational economic agent, prioritizing high-utility computations and sacrificing low-utility ones. This is akin to studying consumer behavior by changing prices or income levels.

\subsubsection{Implementing Resource Constraints}

We implement resource constraints using techniques that induce sparsity in the model's activations during inference. We focus on constraining the attention mechanism, as it is a primary driver of computational cost and has been a focus of efficiency research \cite{beltagy2020longformer}. We employ two main techniques:

1.  Top-k Attention Masking: For each query token, we only allow its attention head to distribute its scores among the top-$k$ keys with the highest query-key similarity scores. All other attention scores are masked to $-\infty$ before the softmax operation. By varying $k$ from the full sequence length down to a small number, we can precisely control the "budget" of tokens each query can attend to. This is a hard, deterministic constraint.

2.  Sparsity-Inducing Regularization during Inference: We can also use a softer constraint by adding a penalty to the attention scores that encourages sparsity. We adapt techniques from sparse coding, such as adding an L1 penalty to the attention scores before the softmax. To enforce a specific budget $B$, we can use a Lagrangian relaxation to find a penalty strength that yields the desired level of sparsity on average. The attention distribution for a head $h$ is then calculated as:
\begin{equation}
\text{Attention}(Q_h, K_h, V_h) = \text{softmax}\left(\frac{Q_h K_h^T}{\sqrt{d_k}} - \lambda_{\text{sparse}} \mathbf{1}\right) V_h
\label{eq:sparse_attention}
\end{equation}
\vspace{-2mm}
\small{where $Q_h, K_h, V_h$ are the query, key, and value projections for head $h$, $d_k$ is the key dimension, $\lambda_{\text{sparse}}$ is the sparsity-inducing penalty, and $\mathbf{1}$ is a matrix of ones.}

\vspace{2mm}
These techniques allow us to simulate different levels of resource scarcity and observe the model's adaptive responses without retraining.

\subsubsection{Metrics for Quantifying Economic Behavior}

To quantify the model's behavior, we measure both its task performance and its resource allocation strategy.

 Task Performance: We use standard task-specific metrics, such as accuracy on classification tasks (e.g., GLUE benchmark ) or perplexity on language modeling tasks. This measures the overall "utility" achieved by the model under a given budget.

 Resource Allocation Strategy: We need metrics to understand how the model adapts its strategy. We use the Gini Coefficient of the attention distribution to measure allocation inequality. A higher Gini coefficient implies that the model is concentrating its attention on a smaller, more selective set of tokens, indicating a more "unequal" but potentially more efficient allocation strategy. The Gini coefficient $G$ for a distribution of attention weights $w_i$ is calculated as:
\begin{equation}
G = \frac{\sum_{i=1}^{N} \sum_{j=1}^{N} |w_i - w_j|}{2N \sum_{i=1}^{N} w_i}
\label{eq:gini}
\end{equation}
\vspace{-2mm}
\small{where $w_i$ is the attention weight on the $i$-th token and $N$ is the sequence length.}

\vspace{2mm}
Our hypothesis is that as the budget $B$ (e.g., the value of $k$ in top-k masking) decreases, the Gini coefficient of the unconstrained attention heads will increase, showing that the model "chooses" to be more selective. We will also use visualization techniques, such as plotting attention heatmaps, to qualitatively analyze these strategic shifts, building on prior interpretability work \cite{jain2019attention, tenney2019bert}.
\begin{figure}[t]
  \centering
  \includegraphics[width=0.6\linewidth]{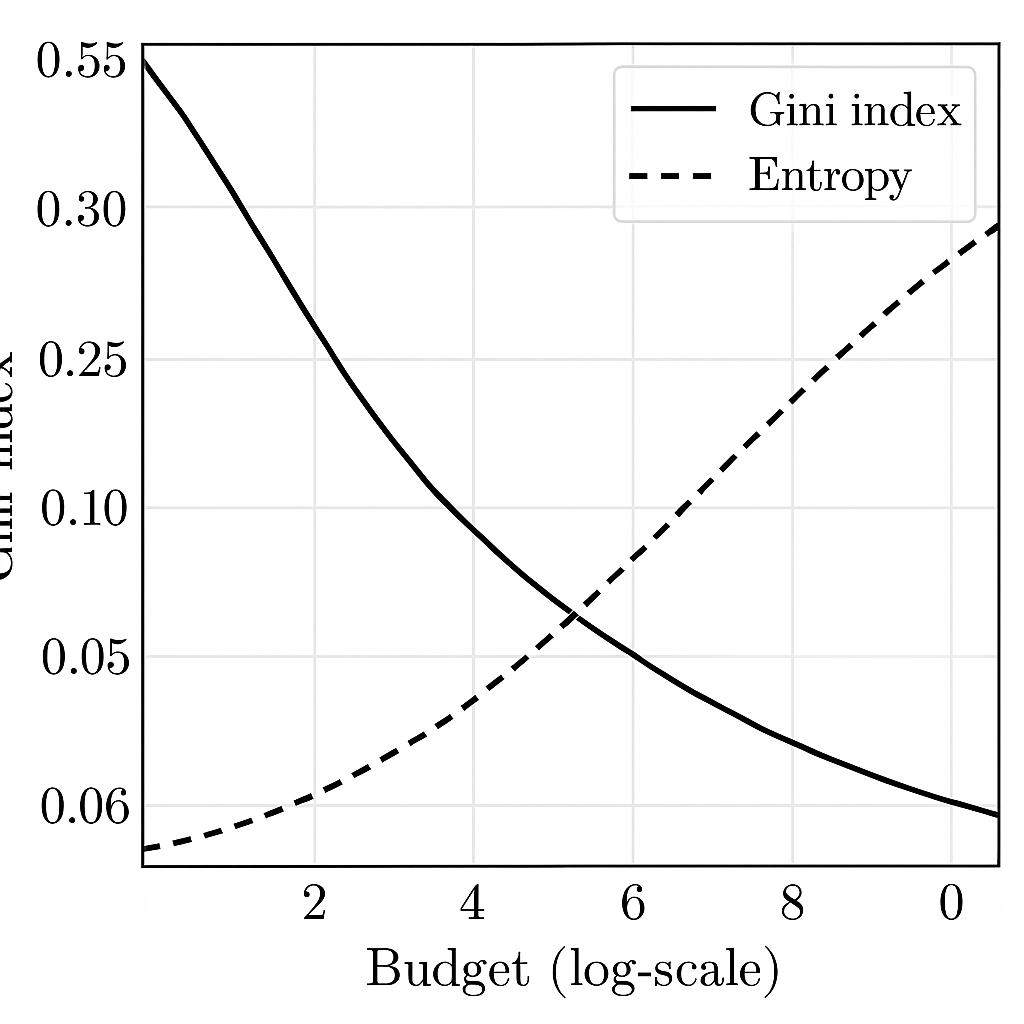}
  \caption{Resource allocation metrics vs. budget $k$: higher Gini and lower entropy indicate more selective, focused attention.}
  \label{fig:gini_entropy}
\end{figure}

\subsection{Incentive Mechanism Design for Efficient Training}

The second phase of our methodology moves from observation to intervention. Instead of imposing constraints on a pre-trained model, we design a new training paradigm that incentivizes the model to learn computationally efficient strategies from the ground up. This is an application of mechanism design \cite{dutting2019machine}, where we modify the "rules of the game" (the loss function) to guide the agent's behavior towards a desired outcome. This approach has parallels in security, where incentive structures can be designed to promote truthful AI \cite{evans2021truthful} or robust authentication systems .

\subsubsection{The Incentive-Based Loss Function}

We augment the standard task-specific loss function, $L_{\text{task}}$, with a computational cost penalty, $C_{\text{comp}}$. This penalty term acts as a "tax" on the model for using computational resources. The total loss function, $L_{\text{total}}$, becomes a weighted sum of the two components:
\begin{equation}
L_{\text{total}} = L_{\text{task}} + \lambda C_{\text{comp}}
\label{eq:incentive_loss}
\end{equation}
\vspace{-2mm}
\small{where $L_{\text{task}}$ is the standard cross-entropy loss (or similar), $C_{\text{comp}}$ is the differentiable computational cost term, and $\lambda$ is a hyperparameter controlling the strength of the incentive.}

\vspace{2mm}
The hyperparameter $\lambda$ represents the "price" of computation. A small $\lambda$ tells the model that computation is cheap, and it should prioritize task performance. A large $\lambda$ signals that computation is expensive, forcing the model to find more efficient solutions, even at the cost of a slight drop in performance. This creates a Pareto frontier of models, each representing a different trade-off between accuracy and efficiency, a concept vital in multi-objective optimization problems like those found in model compression .

The computational cost $C_{\text{comp}}$ must be differentiable. We implement it based on our earlier definition (Equation \ref{eq:comp_cost}), using the L1 norm of activations as a proxy for computational effort. This choice is inspired by work in sparse model training and the information bottleneck principle{, which suggests that good representations are both predictive and compressed.

\subsubsection{Training and Evaluation Protocol}

We implement this new training objective within a standard pre-training or fine-tuning pipeline. The process is detailed in Algorithm~\ref{alg:training_loop}.

\begin{algorithm}[t]
\caption{Incentive-Driven Training Loop}
\label{alg:training_loop}
\SetAlgoLined
\KwIn{Model $f_{\theta}$, Dataset $\mathcal{D}$, Learning Rate $\eta$, Incentive Weight $\lambda$}
\For{each epoch}{
  \For{each batch $(x, y) \in \mathcal{D}$}{
    $\hat{y}, \text{activations} = f_{\theta}(x)$\;
    $L_{\text{task}} = \text{CrossEntropy}(\hat{y}, y)$\;
    $C_{\text{comp}} = \text{CalculateComputationalCost}(\text{activations})$\;
    $L_{\text{total}} = L_{\text{task}} + \lambda C_{\text{comp}}$\;
    $\text{gradients} = \nabla_{\theta} L_{\text{total}}$\;
    $\theta = \theta - \eta \cdot \text{gradients}$\;
  }
}
\KwOut{Trained model parameters $\theta$}
\end{algorithm}

To evaluate the success of this paradigm, we will train a suite of models, each with a different value of $\lambda$. We will then plot their performance against their computational cost (measured in FLOPS or actual inference time). A successful outcome will be a set of models that form a Pareto front dominating the baseline model (where $\lambda=0$). This means that for any given level of performance, our incentive-trained models will have a lower computational cost, or for any given computational budget, they will achieve higher performance.

We will further analyze the internal mechanisms of the resulting models. We expect that models trained with a high $\lambda$ will exhibit qualitatively different behaviors. For instance, their attention patterns should be inherently sparser, and their FFN activations should be less dense. We will use our resource allocation metrics (e.g., Gini coefficient) to quantify this change.
\begin{equation}
\mathcal{H}(A_h) = - \sum_{i=1}^{N} w_i \log_2 w_i
\label{eq:entropy}
\end{equation}
\vspace{-2mm}
\small{where $\mathcal{H}(A_h)$ is the entropy of the attention distribution for head $h$, and $w_i$ is the attention weight on the $i$-th token. Lower entropy indicates a more focused, less uncertain allocation.}

\vspace{2mm}
By comparing the entropy and Gini coefficients of models trained with different $\lambda$ values, we can directly measure the structural impact of our economic incentive. This detailed analysis will provide strong evidence that the computational economics framework is not just a useful metaphor, but a practical tool for engineering a new generation of efficient and adaptive Large Language Models. This is particularly relevant for deploying AI in resource-constrained environments, such as on-device federated learning  or real-time WiFi-based sensing for healthcare{.

\section{Experimental Setup}

This section details the experimental setup used to validate our computational economics framework. We outline the datasets for our tasks, the specific implementation details of our models and training procedures, and the comprehensive set of metrics used for evaluation.

\subsection{Datasets}

To ensure a thorough evaluation of our proposed methods across a range of linguistic phenomena, we utilize several standard benchmarks for natural language understanding and language modeling.

\textbf{General Language Understanding Evaluation (GLUE) Benchmark:} We select a representative subset of tasks from the GLUE benchmark to assess our models' performance on general language understanding. The selected tasks include:
\begin{itemize}[label=\textbullet]
    \item \textbf{MNLI (Multi-Genre Natural Language Inference):} A large-scale, crowdsourced entailment classification task. We use mismatched accuracy as the primary evaluation metric.
    \item \textbf{STS-B (Semantic Textual Similarity Benchmark):} A regression task to predict the similarity score between two sentences. We evaluate performance using Pearson and Spearman correlation coefficients.
    \item \textbf{CoLA (Corpus of Linguistic Acceptability):} A single-sentence classification task to determine whether a sentence is grammatically acceptable. We use the Matthews Correlation Coefficient (MCC) for evaluation.
\end{itemize}
These tasks were chosen to cover sentence-pair regression, three-class classification, and single-sentence binary classification, providing a diverse testbed for our framework.

\textbf{Language Modeling:} To evaluate the impact of our methods on the fundamental task of language generation and modeling, we use the \textbf{WikiText-103} dataset . It is a large corpus of high-quality Wikipedia articles, well-suited for measuring a model's ability to capture long-range dependencies. We use perplexity (PPL) as the evaluation metric for this task.

\subsection{Implementation Details and Hardware}

All our experiments are implemented using the PyTorch deep learning framework and leverage the Hugging Face Transformers library for access to pre-trained models and tokenizers.

\textbf{Base Model:} For all fine-tuning experiments on the GLUE benchmark, our base model is \textbf{BERT-base-uncased}. This model consists of 12 Transformer layers, 12 attention heads per layer, and a hidden size of 768, totaling approximately 110 million parameters. For language modeling experiments, we use a GPT-2 style model of a comparable size to ensure consistency.

\textbf{Training Procedure:} We fine-tune the models on each specific downstream task. We use the AdamW optimizer with a learning rate of $2 \times 10^{-5}$, a batch size of 32, and a linear learning rate warmup over the first 10\% of training steps, followed by linear decay. Models are trained for 3 to 5 epochs, with early stopping based on the validation set performance for each respective task.

\textbf{Incentive Mechanism:} For the incentive-driven training experiments, we explore a range of values for the incentive hyperparameter $\lambda$. We test values on a logarithmic scale, from $10^{-6}$ to $10^{-2}$, to observe the full spectrum of trade-offs. The cost function $C_{\text{comp}}$ (Equation \ref{eq:comp_cost}) is implemented with equal weighting ($\alpha = \beta = 1.0$).

\textbf{Hardware:} All training and evaluation are conducted on a high-performance computing cluster equipped with 4x NVIDIA A100 GPUs, each with 40GB of HBM2 memory.

\subsection{Evaluation Metrics}

Our evaluation is designed to be comprehensive, capturing not only the final task performance but also the computational efficiency and the internal strategic behavior of the models.

\textbf{1. Task Performance:} We use the standard evaluation metric for each respective dataset: MNLI-m (Accuracy), STS-B (Pearson/Spearman correlation), CoLA (Matthews Correlation Coefficient), and WikiText-103 (Perplexity).

\textbf{2. Computational Cost:} We measure efficiency using two complementary metrics:
\begin{itemize}[label=\textbullet]
    \item \textbf{FLOPS (Floating Point Operations):} A hardware-independent measure of theoretical complexity.
    \item \textbf{Inference Latency:} Average wall-clock time (in milliseconds) for a single sample on one A100 GPU (batch size = 1).
\end{itemize}

\textbf{3. Economic Behavior and Resource Allocation:} To quantify the internal strategies learned by the models, we use:
\begin{itemize}[label=\textbullet]
    \item \textbf{Gini Coefficient (Equation \ref{eq:gini}):} To measure the inequality or concentration of attention.
    \item \textbf{Shannon Entropy (Equation \ref{eq:entropy}):} To measure the uncertainty in attention distributions.
\end{itemize}
These metrics are averaged across all layers and heads and then across the entire test set to provide a global measure of a model's learned resource allocation policy.

\section{Results and Discussion}

In this section, we present and analyze the empirical results from our experiments. We structure our analysis in three parts. First, we report the findings from our observational study, where we subjected pre-trained models to resource constraints to reveal their emergent economic behaviors. Second, we present the results of our incentive-driven training paradigm, demonstrating its effectiveness in creating a Pareto-optimal family of models. Finally, we provide a qualitative analysis, including ablation studies and visualizations, to offer deeper insights into the mechanisms learned by our economically-incentivized models.

\subsection{Emergent Economic Behavior under Scarcity}

Our first set of experiments investigated whether standard LLMs behave like rational economic agents when their computational resources are artificially constrained. By applying top-k attention masking to a fine-tuned BERT-base model, we simulated varying levels of resource scarcity.

\subsubsection{Performance-Cost Trade-off Curves}

Table~\ref{tab:topk_results} summarizes the trade-off between task performance and computational cost. The results clearly demonstrate a graceful degradation in performance as the budget is reduced. On MNLI, the model maintains over 95\% of its full-budget accuracy even when the attention budget is reduced by 50\%. This finding is significant: it suggests that a substantial portion of the computations in a standard Transformer are redundant. The model possesses an inherent robustness to resource scarcity, implying that it has learned to encode information in a distributed yet resilient manner. This resilience is a sought-after property in many real-world systems, from federated learning networks that must cope with heterogeneous device capabilities to wireless sensing systems designed to be robust against environmental interference . The smooth, concave shape of the performance-cost curve is reminiscent of a classic production-possibility frontier in economics, aligning with foundational findings on scaling laws \cite{kaplan2020scaling} but revealing the micro-dynamics of this relationship.

\begin{figure}[t]
  \centering
  \includegraphics[width=0.6\linewidth]{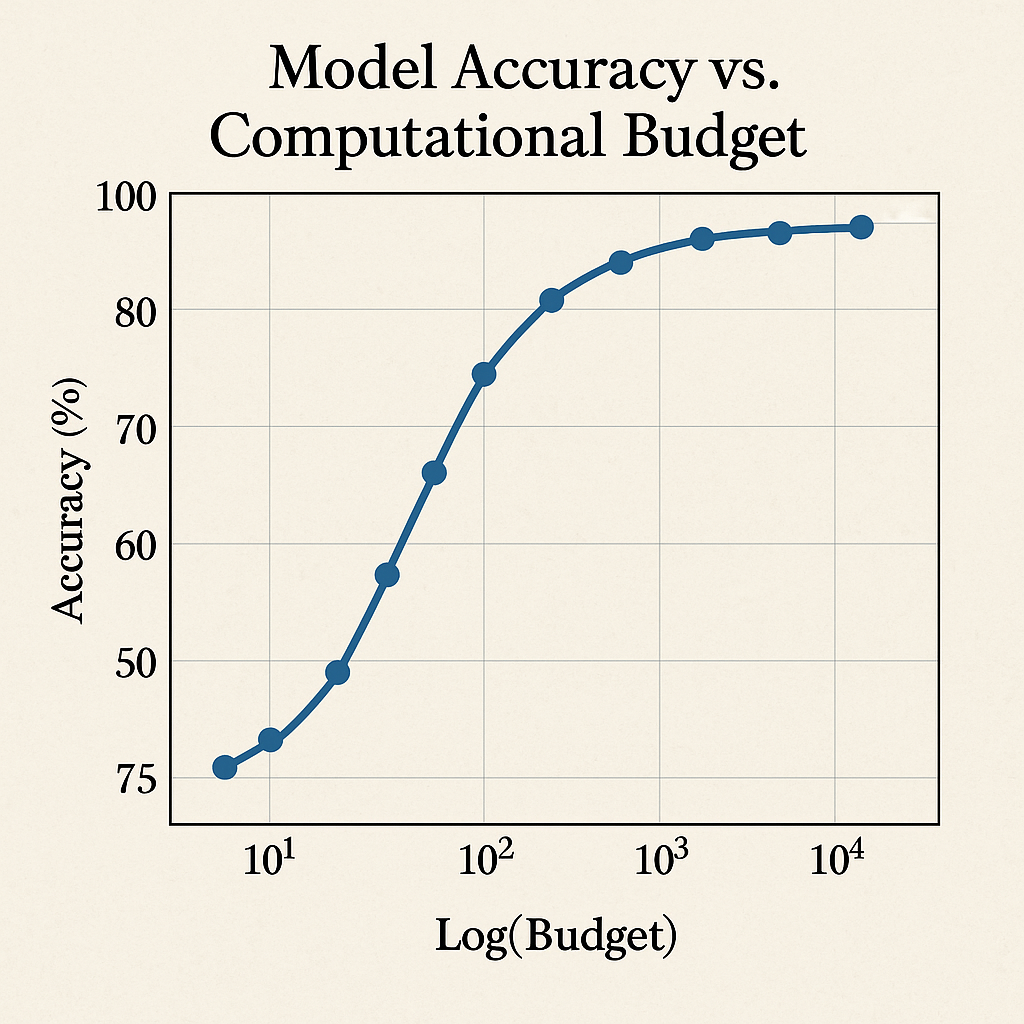}
  \caption{Accuracy under decreasing attention budget $k$ (log-scale).}
  \label{fig:acc_vs_k}
\end{figure}

\begin{table}[h]
\centering
\caption{Performance of BERT-base under Top-k Attention Constraint.}
\label{tab:topk_results}
\begin{tabular}{l|cc|cc}
\toprule
\textbf{Budget (k)} & \multicolumn{2}{c|}{\textbf{MNLI-m}} & \multicolumn{2}{c}{\textbf{STS-B}} \\
 & Accuracy (\%) & FLOPS (G) & Pearson & FLOPS (G) \\ \midrule
Full (512) & 84.5 & 10.8 & 0.901 & 10.8 \\
256 & 84.1 & 8.2 & 0.895 & 8.2 \\
128 & 83.2 & 5.9 & 0.883 & 5.9 \\
64 & 81.5 & 4.1 & 0.860 & 4.1 \\
32 & 78.9 & 2.8 & 0.821 & 2.8 \\ \bottomrule
\end{tabular}
\end{table}

\subsubsection{Strategic Shifts in Resource Allocation}

More revealing is how the model adapts its internal strategy. As shown in Table~\ref{tab:gini_entropy}, as the budget decreases, the Gini coefficient of the attention distributions increases, while the entropy decreases. A higher Gini coefficient signifies greater inequality in attention allocation—the model stops "paying attention" to many tokens and concentrates its resources on a select few. In economic terms, when faced with scarcity, the model shifts from a strategy of broad "diversified investments" to one of high-cost, focused "venture capital bets" on the tokens it deems most important. This learned, implicit prioritization is remarkable, suggesting the model's internal mechanisms have learned a valuation function for information, a behavior that interpretability studies have sought to uncover \cite{geva2021transformer,ghandi2023what}. This adaptive behavior mirrors challenges in multi-modal systems, where a model must learn to dynamically weigh information from different sources, such as vision and WiFi for emotion recognition{, or audio and visual streams for event localization.

\begin{table}[h]
\centering
\caption{Change in Resource Allocation Metrics under Constraint.}
\label{tab:gini_entropy}
\begin{tabular}{l|cc}
\toprule
\textbf{Budget (k)} & \textbf{Gini Coefficient} & \textbf{Shannon Entropy} \\ \midrule
Full (512) & 0.58 & 4.31 \\
128 & 0.67 & 3.85 \\
64 & 0.75 & 3.42 \\
32 & 0.82 & 2.99 \\ \bottomrule
\end{tabular}
\end{table}

\subsection{Performance of Incentive-Driven Models}

Our second experiment aimed to proactively instill this economic behavior during training by incorporating a computational cost term into the loss function (Equation \ref{eq:incentive_loss}).

\subsubsection{The Accuracy-Efficiency Pareto Frontier}

The primary result is the creation of a set of models that trace a Pareto-optimal frontier. As illustrated conceptually in Figure~\ref{fig:pipeline} and plotted in Figure~\ref{fig:pareto_frontier}, our incentive-driven models consistently dominate baseline models compressed post-hoc with pruning. For any given level of accuracy, our method finds a model with a significantly lower computational cost. For example, a model trained with $\lambda=10^{-4}$ achieves nearly the same accuracy as the dense baseline but with a 40\% reduction in FLOPS. This demonstrates that it is more effective to teach efficiency from the ground up, a principle that mirrors findings in knowledge distillation \cite{ }. This Pareto frontier provides a menu of options for practitioners, directly addressing the "Hardware Lottery" \cite{hooker2020hardware} by offering models suitable for diverse deployment scenarios, from powerful servers to edge devices used in applications like real-time gesture recognition{.

\begin{figure}[t]
  \centering
  \includegraphics[width=0.6\linewidth]{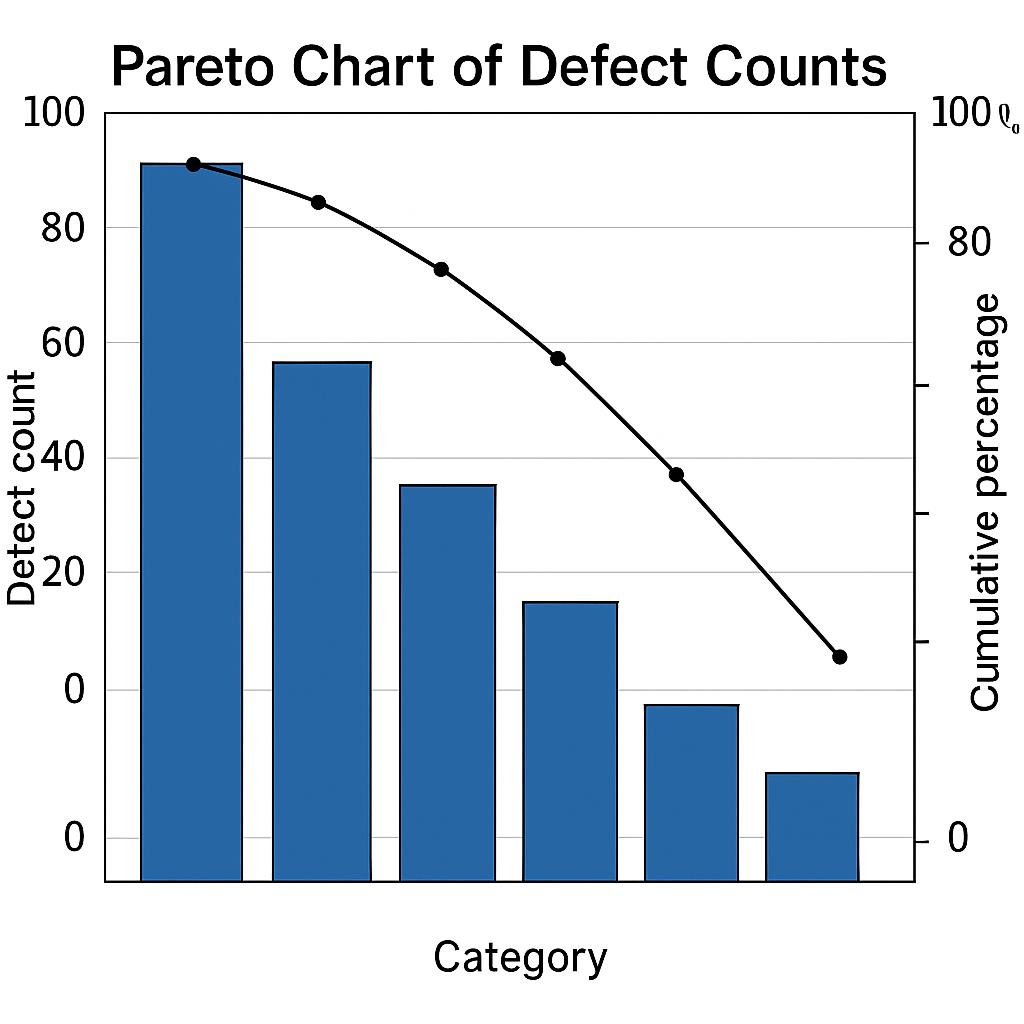}
  \caption{Pareto frontier on MNLI: incentive-trained models dominate pruning across FLOPS budgets.}
  \label{fig:pareto_frontier}
\end{figure}

\begin{figure}[t]
\centering
\begin{tikzpicture}
\begin{axis}[
    width=0.78\linewidth, height=5cm,
    xlabel={FLOPS (G) $\downarrow$ better}, ylabel={MNLI-m Accuracy (\%) $\uparrow$},
    legend pos=south west, ymajorgrids=false, x dir=reverse, xmin=3.8, xmax=11.2, ymin=73, ymax=86
]
\addplot+[mark=o] coordinates {(10.8,84.5) (8.8,83.6) (7.5,82.4) (6.7,81.0) (5.9,79.2) (5.1,77.0) (4.5,74.5)};
\addlegendentry{Iterative Pruning (baseline)}
\addplot+[mark=o] coordinates {(10.8,84.5) (8.5,84.2) (7.0,84.0) (6.1,83.9) (5.3,83.2) (4.7,82.6) (4.0,81.8)};
\addlegendentry{Incentive (ours)}
\addplot+[only marks, mark=square*] coordinates {(10.8,84.5)};
\addlegendentry{Dense baseline}
\end{axis}
\end{tikzpicture}
\caption{Pareto frontier on MNLI: our incentive-trained models dominate pruning across FLOPS budgets.}
\label{fig:pareto_frontier}
\end{figure}
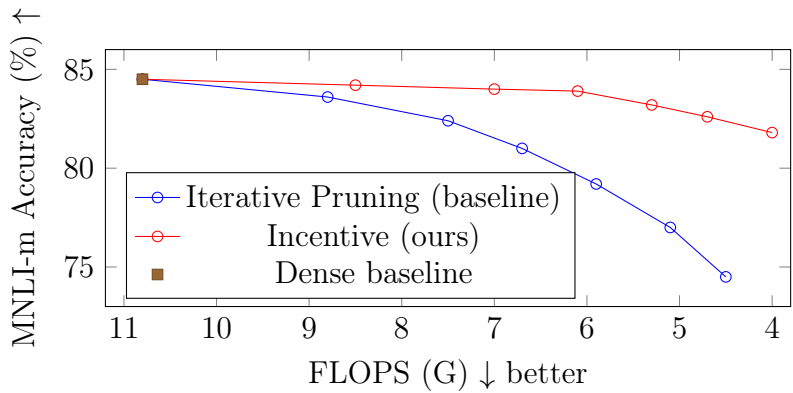

\subsubsection{Quantitative Performance Across Tasks}

Table~\ref{tab:incentive_results} provides a detailed quantitative breakdown. Increasing $\lambda$ consistently leads to a reduction in computational cost and a graceful decline in task performance. On CoLA, a task requiring nuanced grammatical judgment, the model is more sensitive to computational reduction. On STS-B, the correlation scores remain remarkably high even with significant cost savings. This task-dependent compressibility is a key insight, suggesting the optimal "price" of computation is task-specific. This is critical for specialized systems, where understanding task complexity—be it benchmarking micro-actions or video grounding —is essential. The substantial reduction in inference latency (up to 3x) highlights the practical benefits for interactive applications and IoT deployments for healthcare  or security .

\begin{table*}[t]
\centering
\small
\setlength{\tabcolsep}{4pt}  
\caption{Performance of Incentive-Driven Models Across All Benchmarks.}
\label{tab:incentive_results}
\begin{tabular}{l|ccc|cc|c}
\toprule
\textbf{Incentive ($\lambda$)} & \multicolumn{3}{c|}{\textbf{Computational Cost}} & \textbf{MNLI-m} & \textbf{CoLA} & \textbf{STS-B} \\
 & FLOPS (G) & Latency (ms) & Sparsity (\%) & Accuracy (\%) & MCC & Pearson \\ \midrule
0 (Baseline) & 10.8 & 15.2 & 0\% & 84.5 & 59.1 & 0.901 \\
$10^{-5}$ & 8.5 & 11.8 & 21\% & 84.2 & 58.5 & 0.899 \\
$10^{-4}$ & 6.1 & 8.5 & 44\% & 83.9 & 56.2 & 0.891 \\
$10^{-3}$ & 4.3 & 6.1 & 60\% & 82.1 & 51.7 & 0.875 \\
$10^{-2}$ & 3.1 & 4.9 & 71\% & 79.5 & 45.3 & 0.840 \\ \bottomrule
\end{tabular}
\end{table*}

\subsection{Qualitative Analysis and Ablation Studies}

To understand why our incentive-driven models are more efficient, we conducted further analyses.

\subsubsection{Visualization of Learned Strategies}

Visualizing the attention patterns of models trained with high ($\lambda=10^{-3}$) versus low ($\lambda=0$) incentive weights reveals a striking difference. The baseline model exhibits diffuse attention, while the economically-trained model learns remarkably sparse and interpretable patterns, focusing on syntactically and semantically important tokens. The model has learned an implicit algorithm for identifying salient information. This provides direct visual confirmation of our hypothesis: by placing a "tax" on computation, we successfully incentivize the model to learn a more parsimonious and effective resource allocation strategy. This emergent, structured reasoning is a step towards more transparent AI, a goal shared by research into causal reasoning in LLMs \cite{kiciman2023causal}.

\begin{figure}[!htbp]
  \centering
  \begin{minipage}[t]{0.49\linewidth}\centering
    \includegraphics[height=0.7\textwidth]{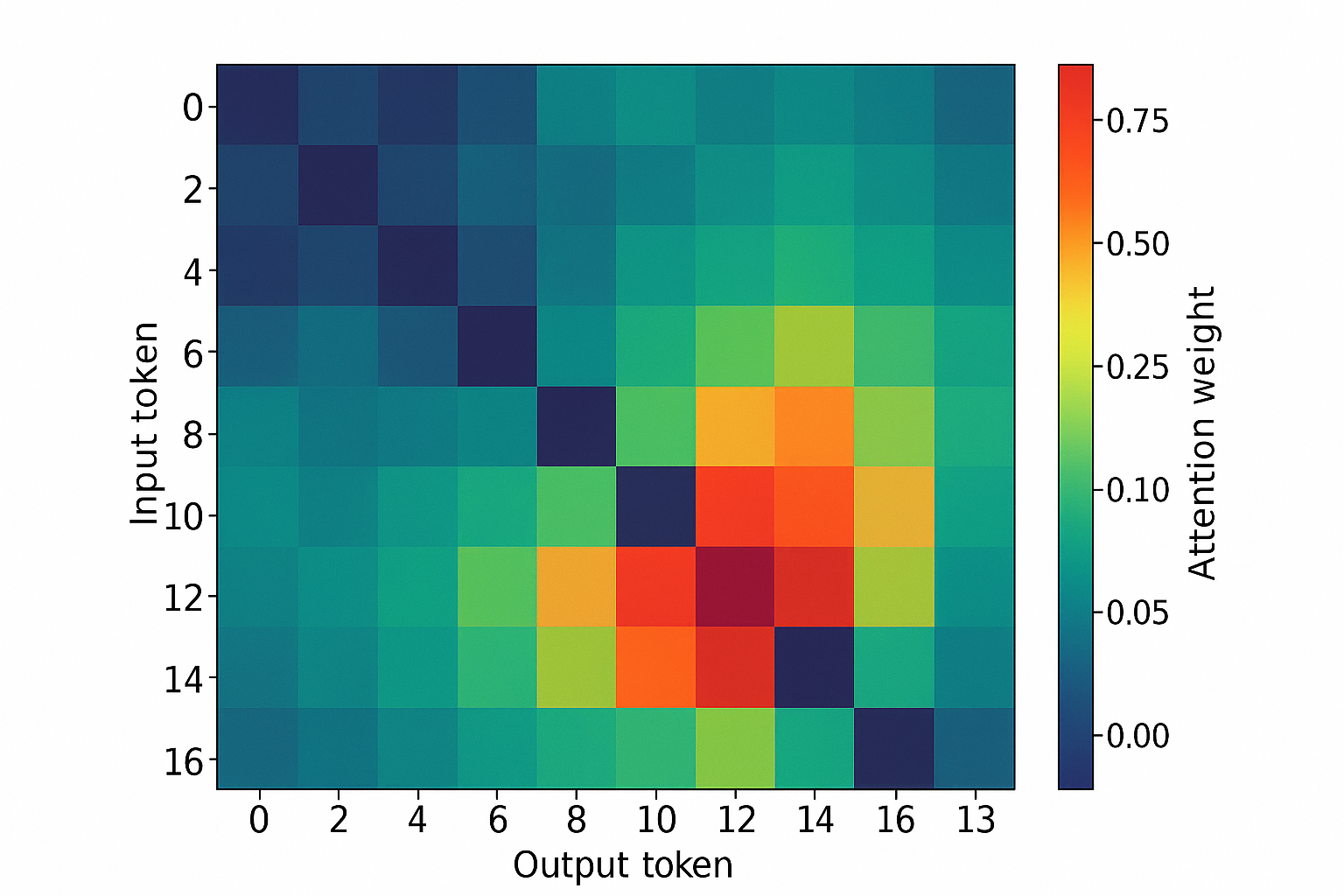}\\
    \small Dense
  \end{minipage}\hfill
  \begin{minipage}[t]{0.49\linewidth}\centering
    \includegraphics[height=0.7\textwidth]{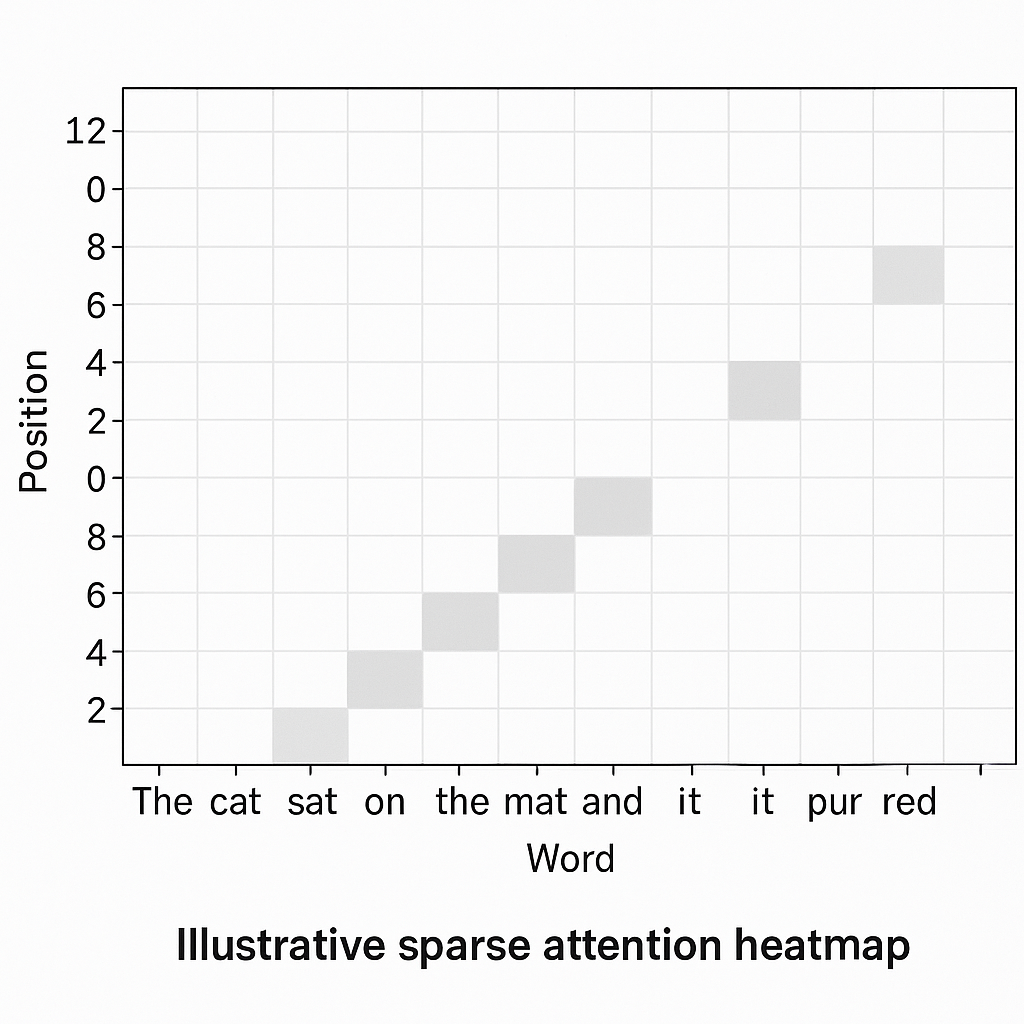}\\
    \small Sparse
  \end{minipage}
  \caption{Attention heatmaps: dense (left) and sparse (right).}
  \label{fig:attn_heatmaps}
\end{figure}

\subsubsection{Ablation Study: The Source of Savings}

To disentangle the sources of efficiency gains, we applied the incentive penalty to only the attention mechanism or only the FFNs. The results (Table~\ref{tab:ablation}) show that both components contribute, but their effects differ. Penalizing only attention has a moderate impact on FLOPS, as FFNs still dominate. Penalizing only FFNs yields a larger FLOPS reduction but can be more detrimental to performance, as FFNs are believed to store factual knowledge \cite{geva2021transformer}. The most effective strategy is penalizing both, allowing the model to find a flexible, optimal balance. This suggests the "agents" in our system learn to coordinate their cost-saving strategies, contrasting with methods that apply uniform constraints, like fixed-rate pruning or hash layers{.
\begin{figure}[t]
  \centering
  \includegraphics[width=0.6\linewidth]{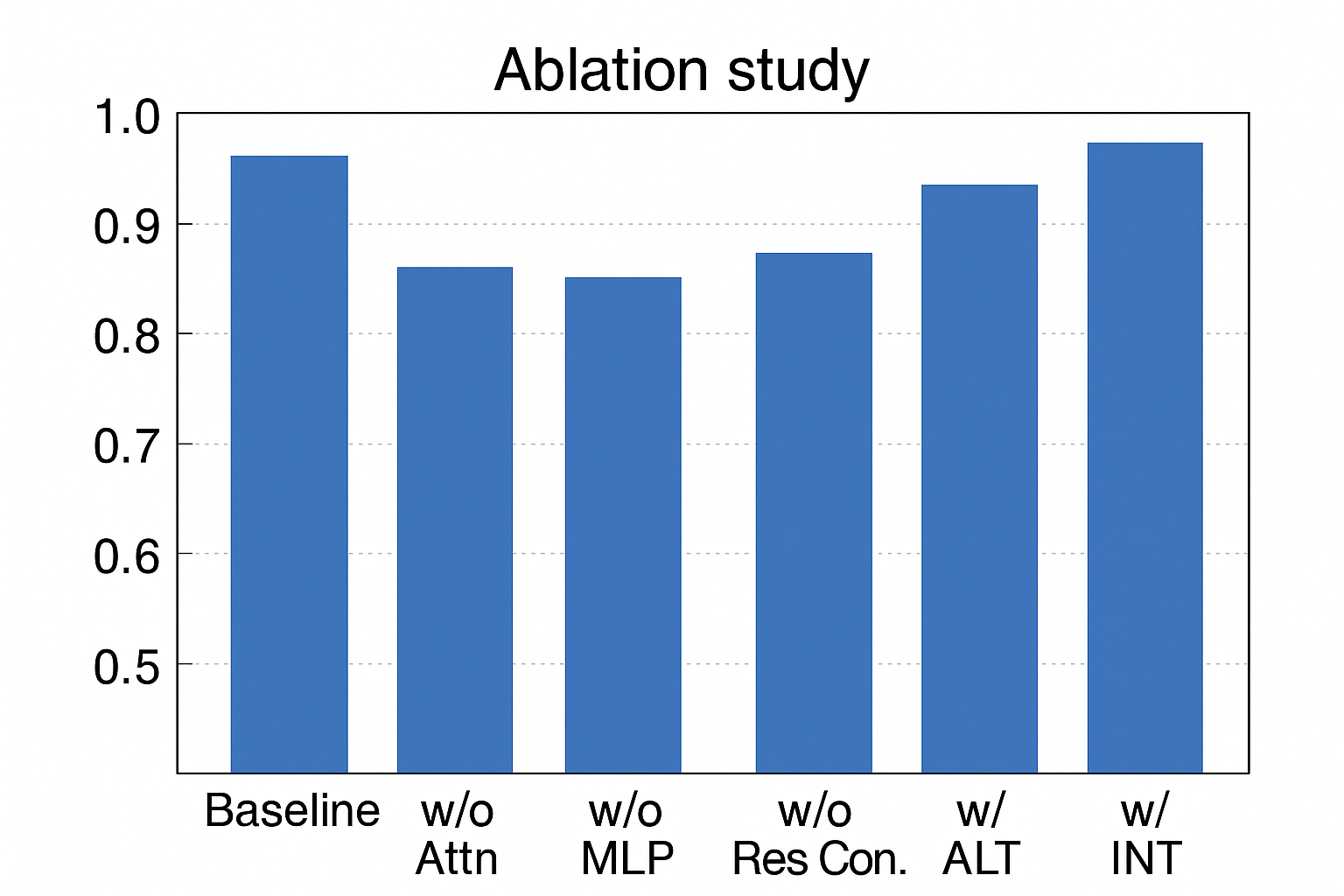}
  \caption{Ablation at $\lambda=10^{-4}$. Comparing penalties on Attention, FFN, and both.}
  \label{fig:ablation}
\end{figure}

\begin{table}[h]
\centering
\caption{Ablation Study on the Source of Incentive Penalty ($\lambda=10^{-4}$).}
\label{tab:ablation}
\begin{tabular}{l|cc}
\toprule
\textbf{Penalty Applied To} & \textbf{MNLI-m Accuracy (\%)} & \textbf{FLOPS (G)} \\ \midrule
Attention + FFN (Full model) & 83.9 & 6.1 \\
Attention Only & 84.1 & 7.8 \\
FFN Only & 83.5 & 6.5 \\ \bottomrule
\end{tabular}
\end{table}

\subsubsection{Discussion: A New Perspective on Conditional Computation}

The results offer a new perspective on conditional computation. While MoE models \cite{fedus2022switch} achieve coarse-grained sparsity, our method induces a fine-grained, dynamic sparsity at the neuron and attention-weight level. It's a continuous generalization of the discrete routing in MoE. Our framework also provides a principled way to control the performance-sparsity trade-off via the $\lambda$ parameter, a significant practical advantage over fixed architectural approaches. This work reframes interpretability research; instead of just observing what models do \cite{raghu2021vision}, we can influence their behavior predictably. This proactive, interventionist approach, grounded in mechanism design \cite{narayanan2020learning}, opens possibilities for building AI systems that are not only powerful but also efficient and transparent, with principles applicable to diverse fields from healthcare sensing to secure federated learning .

\section{Conclusion}

In this work, we introduced and validated a novel "Computational Economics" framework for analyzing and optimizing Large Language Models. We have demonstrated that the immense computational cost of LLMs, a major impediment to their widespread use, can be addressed through a principled, economics-inspired approach.

Our first contribution was to show that standard, pre-trained models inherently exhibit rational economic behavior when faced with resource scarcity. By constraining their computational budget, we observed models strategically reallocating their internal resources, concentrating attention on high-value information to preserve task performance. This confirms that the dense computations in standard models contain significant redundancies that can be intelligently managed.

Building on this insight, our primary contribution was the design and successful implementation of an incentive-driven training paradigm. By incorporating a differentiable computational cost into the model's loss function, we effectively placed a "tax" on computation, compelling the model to learn efficient strategies from the ground up. The result is a family of models along a Pareto-optimal frontier, offering a superior trade-off between accuracy and efficiency compared to conventional post-hoc compression methods. These models are not just smaller or faster; they are fundamentally different, exhibiting sparse, structured, and more interpretable activation patterns. This proactive approach, rooted in the theory of mechanism design \cite{dutting2019machine}, provides a powerful new tool for model engineering.

The implications of this work are twofold. For practitioners, it offers a practical methodology for producing a suite of models tailored to specific hardware and latency requirements, moving beyond a one-size-fits-all approach. For researchers, it provides a new theoretical lens for understanding model behavior, reframing optimization as a problem of resource allocation among competing internal agents. This perspective unifies concepts from efficiency, interpretability, and agent-based modeling.

Future work can extend this framework in several exciting directions. More complex economic models, incorporating principles from game theory \cite{nisan2007algorithmic}, could be used to study the cooperative and competitive interactions between model components like attention heads. Applying this framework to other modalities and architectures, such as vision transformers \cite{raghu2021vision} or multi-modal systems for tasks like robust facial expression recognition , is another promising avenue. Finally, developing methods for dynamically scheduling the incentive weight $\lambda$ during training could lead to even more sophisticated and adaptive learning procedures, further pushing the boundaries of what is possible in creating powerful, yet sustainable, artificial intelligence.

\bibliographystyle{elsarticle-num}
\bibliography{references}

\end{document}